# An MDP-Based Recommender System

**Guy Shani**
Department of Computer Science
Ben-Gurion University
Beer-Sheva 84105, Israel
shanigu@cs.bgu.ac.il

**Ronen I. Brafman**
Department of Computer Science
Ben-Gurion University
Beer-Sheva 84105, Israel
brafman@cs.bgu.ac.il

**David Heckerman**
Microsoft Research
One Microsoft Way
Redmond, WA 98052
heckerma@microsoft.com

## Abstract

Typical Recommender systems adopt a static view of the recommendation process and treat it as a prediction problem. We argue that it is more appropriate to view the problem of generating recommendations as a sequential decision problem and, consequently, that Markov decision processes (MDP) provide a more appropriate model for Recommender systems. MDPs introduce two benefits: they take into account the long-term effects of each recommendation, and they take into account the expected value of each recommendation. To succeed in practice, an MDP-based Recommender system must employ a strong initial model; and the bulk of this paper is concerned with the generation of such a model. In particular, we suggest the use of an $n$-gram predictive model for generating the initial MDP. Our $n$-gram model induces a Markov-chain model of user behavior whose predictive accuracy is greater than that of existing predictive models. We describe our predictive model in detail and evaluate its performance on real data. In addition, we show how the model can be used in an MDP-based Recommender system.

## 1 INTRODUCTION

In many markets, consumers are faced with a wealth of products and information from which they can choose. To alleviate this problem, many web sites attempt to help users by incorporating a *Recommender system* (Schafer, Konstan, & Riedl, 1999) that provides users with a list of items and/or web-pages that are likely to interest them. Once the user makes her choice, a new list of recommended items is presented. Thus, the recommendation process is a sequential process. Moreover, in many domains, user choices are sequential in nature, e.g., we buy a book by the author of a recent book we liked.

The sequential nature of the recommendation process was noticed in the past (Zimdars, Chickering, & Meek, 2001). Taking this idea one step farther, we suggest that recommending is not simply a sequential prediction problem, but rather, a sequential decision problem. At each point the Recommender system makes a decision: which recommendation to issue. This decision should be optimized taking into account the sequential process involved and the optimization criteria suitable for the Recommender system. Thus, we suggest the use of Markov Decision Processes (MDP) (e.g., Puterman, 1994), a well known stochastic model of sequential decisions. With this view in mind, a more sophisticated approach to Recommender systems emerges. First, one can take into account the utility of a particular recommendation – e.g., we might want to recommend a product that has a slightly lower probability of being bought, but generates higher profits. Second, we might suggest an item whose immediate reward is lower, but leads to more likely or more profitable rewards in the future. These considerations are taken into account automatically by any good or optimal policy generated for an MDP model of the recommendation process. For instance, consider a site selling electronic appliances faced with the option to suggest a video camera with a success probability of 0.5, or a VCR with a probability of 0.6. The site may choose the camera, which is less profitable, because the camera has accessories that are likely to be purchased, where as the VCR does not. If a video-game console is another, less likely option, the large profit from the likely future event of selling game cartridges may tip the balance toward this latter choice.

Keeping these benefits in mind, we suggest an approach for the construction of an MDP-based Recommender system. The first, and most crucial step in the construction of our model is the generation of a powerful predictive model. We believe that in any real environment, it is essential to start with a powerful initial model for the MDP—commercial sites will not accept a model that generates poor recommendations. Use of standard reinforcement learning techniques for MDPs would perform poorly at first due to the amount of needed explorations before converging into a near-optimal policy. Thus, the bulk of this paper is concerned with defining a predictive model that provides accurate initial recommendations.

The predictive model we describe is motivated by our sequential view of the recommendation process, but constitutes an independent contribution. The model can be thought of as an $n$-gram model (e.g., Chen & Goodman, 1998) or, equivalently, a (first-order) Markov chain in which states correspond to sequences of events. In this paper, we emphasize the latter interpretation due to its natural relationship with an MDP. We note that Su, Yang, and Zhang (2000) have described the use of simple $n$-gram models for predicting web pages. Their methods, however, yield poor performance on our data.

Validating our MDP approach is not simple. Most Recommender systems, such as dependency networks (Heckerman, Chickering, Meek, Rounthwaite, & Kadie, 2000), are tested on historical data for their predictive accuracy. That is, the system is trained using historical data from sites that do not provide recommendations, and tested to see whether the recommendations conform to actual user behavior. This test provides some indication of the system's abilities, but it does not test how user behavior is influenced by the system's suggestions or what percentage of recommendations are accepted by users. To obtain this data, one must employ the system in a real site with real users, and compare the performance of this site with and without the system (or with this and other systems). We also note that testing our MDP model using standard techniques would yield worse results than using the Markov- chain model since the MDP would recommend items by their expected value to the site, rather than by their probability of being interesting to the user. For this reason, testing the MDP model on a test bed that doesn't allow interactions with users is useless. This issue has been a problem for past work as well as for this work – it is difficult to convince commercial sites to use experimental systems, and nearly impossible to convince them to compare two different systems. Thus, like most previous work, we evaluate only the predictive component of our model.

The paper is structured as follows. In Section 2, we describe our predictive model—a model that predicts the user's next selection based on her previous ones. In Section 3, we formalize the recommendation problem as an MDP, and show how the predictive model can be used to initialize the MDP. In Section 4, we evaluate the performance of our predictive algorithm on real data. In Section 5, we conclude the paper with a summary and discussion of future work.

## 2 PREDICTIVE MODEL

In this section, we describe our model for predicting the user's next selection based on her previous selections. As we have mentioned, the model can be described as a (first-order) Markov chain. A Markov chain (MC) consists of a set of states, a stochastic transition function denoting the probability distribution over states at sequence point $t$, given the state at sequence point $t-1$, and an initial probability distribution over states. Our transition function is based on the $n$-gram model class, which we now describe.

### 2.1 $N$-GRAM MODELS

$N$-gram models originate in the field of language modeling. They are used to predict the next word in a sentence given the last $n-1$ words. In the simplest form of the model, probabilities for the next word are estimated via maximum likelihood; and many methods exist for improving this simple approach including skipping, clustering, and smoothing. Skipping assumes that the probability of the next word $x_i$ depends on words other than just the previous $n-1$. A separate model is built using skipping and then combined with the standard $n$-gram model. Clustering is an approach that groups some states together for purposes of predicting next states. Such grouping helps to address the problem of data sparsity. Smoothing is a general name for methods that adjust the estimates of probabilities to achieve higher accuracy by adjusting zero or low probabilities upward. One type of smoothing is finite mixture modeling, which combines multiple models via a convex combination. In particular, given $k$ component models for $x_i$ given a prior sequence $X$ — $p_{M_1}(x_i|X), \ldots, p_{M_k}(x_i|X)$ — we can define the $k$-component mixture model $p(x_i|X) = \pi_1 p_{M_1}(x_i|X) + \cdots + \pi_k p_{M_k}(x_i|X)$, where $\sum_{i=1}^{k} \pi_i = 1$ are its mixture weights. Details of these and other methods are given in (e.g., Chen & Goodman, 1998).

### 2.2 AN $N$-GRAM MOTIVATED MARKOV-CHAIN PREDICTION MODEL

**States.** The states in our MC model represent the relevant information that we have about the user. This information corresponds to previous choices made by users in the form of a set of ordered sequences of selections. We ignore data such as age or gender, although it could be beneficial, and it can be easily incorporated into our model. Thus, the set of states contains all possible sequences of user selections. Of course, this formulation leads to an unmanageable state space with the usual associated problems—data sparsity and MDP solution complexity. To reduce the size of the state space, we consider only sequences of at most $k$ items, for some relatively small value of $k$. We note that this approach is consistent with the intuition that the near history (e.g. the current user session) often is more relevant than selections made less recently (e.g. in past user sessions). These sequences are represented as vectors of size $k$. In particular, we use $\langle x_1, \ldots, x_k \rangle$ to denote the state in which the user's last $k$ selected items were $x_1, \ldots, x_k$. Selection sequences with $l < k$ items are transformed into a vector in which $x_1$ through $x_{k-l}$ have the value *missing*. The initial state in the Markov chain is the state in which every entry has the value *missing*. Note that this framework also accommodates systems that collect explicit rather than implicit ratings. We can either map such ratings into the states "preferred" and "not preferred", or enrich the state space with the rating values.

Besides addressing sparsity and computational complexity, the restriction to $k$ items seems sensible for many commercial domains where distant history has little or no impact on user choices. In our experiments, we used values of $k$ ranging from 1 to 5. In our description below, we typically fix $k$ to 3. Note that $k$ in the MC formulation corresponds to $n - 1$ in the $n$-gram formulation.

**The Transition Function.** The transition function for our Markov chain describes the probability that a user whose $k$ recent selections were $x_1, \ldots, x_k$ will select the item $x'$ next. This is denoted $tr_{MC}(\langle x_1, \ldots, x_k \rangle, \langle x_2, \ldots, x_k, x' \rangle)$. Initially, this transition function is unknown to us; and we would like to estimate it based on user data. As mentioned, a maximum-likelihood estimate can be used. This model, however, still suffers from the problem of data sparsity (e.g., Sarwar, Karypis, Konstan, & Riedl, 2000b) and performs poorly in practice. In the next section, we describe several techniques for improving the estimate.

## 2.3 IMPROVEMENTS TO THE PREDICTION MODEL

We experimented with several enhancements to the maximum-likelihood $n$-gram model using data different from that used in our formal evaluation. The improvements described and used here are those that were found to work well.

One enhancement is a form of *skipping* (Chen & Goodman, 1998), and is based on the observation that the occurrence of the sequence $x_1, x_2, x_3$ lends some likelihood to the sequence $x_1, x_3$. That is, if a person bought $x_1, x_2, x_3$, then it is likely that someone will buy $x_3$ after $x_1$. The particular skipping model that we found to work well is a simple additive model. First, each state transition is initialized to the number of observed transitions in the data. Then, given a user sequence $x_1, x_2, ..., x_n$, we add the fractional count $1/2^{(j-(i+3))}$ to the transition from $\langle x_i, x_{i+1}, x_{i+2} \rangle$ to $\langle x_{i+1}, x_{i+2}, x_j \rangle$, for all $i + 3 < j \leq n$. This fractional count corresponds to a diminishing probability of skipping a large number of transactions in the sequence. Finally, the counts are normalized to obtain the transition probabilities:

$$tr_{MC}(s, s') = \frac{count(s, s')}{\sum_{s'} count(s, s')}$$

where $count(s, s')$ is the (fractional) count associated with the transition from $s$ to $s'$.

A second enhancement is a form of clustering that we have not found in the literature. Motivated by properties of our domain, the approach exploits similarity of sequences. For example, the state $\langle x, y, z \rangle$ and the state $\langle w, y, z \rangle$ are similar because some of the items appearing in the former appear in the latter as well. The essence of our approach is that the likelihood of transition from $s$ to $s'$ is influenced by occurrences from $t$ to $s'$, where $s$ and $t$ are similar. In particular, we define the similarity of states $s_i$ and $s_j$ to be

$$sim(s_i, s_j) = \sum_{m=1}^{k} \delta(s_i^m, s_j^m) \cdot (m + 1)$$

where $\delta(\cdot, \cdot)$ is the Kroneker delta function and $s_i^m$ is the $m$th item in state $s_i$. As will become clear shortly, this similarity is arbitrary up to a constant. In addition, we define the *similarity count* from state $s$ to $s'$ to be

$$simcount(s, s') = \sum_{s_i} sim(s, s_i) \cdot tr_{MC}^{old}(s_i, s')$$

where $tr_{MC}^{old}(s_i, s')$ is the original transition function (with or without skipping). The new transition probability from $s'$ to $s$ is then given by

$$tr_{MC}(s, s') = \frac{1}{2} tr_{MC}^{old}(s, s') + \frac{1}{2} \frac{simcount(s, s')}{\sum_{s'} simcount(s, s')} \tag{1}$$

A third enhancement is the use of finite mixture modeling.[1] Similar methods are used in $n$-gram models, where—for example—a trigram, a bigram, and a unigram are combined into a single model. Our mixture model is motivated by the fact that larger values of $k$ lead to states that are more informative whereas smaller values of $k$ lead to states on which we have more statistics. To balance these conflicting properties we mix $k$ models, where the $i$th model looks at the last $i$ transactions. Thus, for $k = 3$, we mix three models that predict the next transaction based on the last transaction, the last two transactions, and the last three transactions. In general, we can learn mixture weights from data. We can even allow the mixture weights to depend on the given case (and informal experiments on our data suggest that such context-specificity would improve predictive accuracy). Nonetheless, for simplicity, we use $\pi_1 = \cdots = \pi_k = 1/k$ in our experiments. Because our primary model is based on the $k$ last items, the generation of the models for smaller values entail little computational overhead.

---

[1] Note that Equation 1 is also a simple mixture model.

# 3  AN MDP-BASED RECOMMENDATION STRATEGY

As we have discussed, our ultimate goal is to construct a Recommender system—a system that chooses a link, product, or other item to recommend to the user at all times. In this section, we describe how such a system can be based on an MDP.

## 3.1  MARKOV DECISION PROCESSES (MDPs)

An MDP is a model for sequential stochastic decision problems. It is a four-tuple: $\langle S, A, R, tr \rangle$, where $S$ is a set of states, $A$ is a set of actions, $R$ is a reward function, and $tr$ is the state-transition function.

A state $s \in S$ encapsulates all the relevant information about the state of the world. The actions trigger state changes, and the effect of the actions on the states is captured by the transition function. The transition function assigns a probability distribution to every (state, action) pair. Thus, $tr(s, a, s')$ is the probability of making a transition from state $s$ to state $s'$ when $a$ is performed. Finally, the reward function assigns a real value to each (state, action) pair which describes the immediate reward (or cost) of executing this action in that state. Often, the reward is only a function of the state, and thus, is a measure of the desirability of reaching each state.

In an MDP, the decision-maker's goal is to behave so that some function of its reward stream is maximized – typically the average or discounted average reward. An optimal solution to the MDP is such a maximizing behavior. Formally, a stationary policy for an MDP is a mapping from states to actions, specifying which action we should perform at each state. A history-dependent policy associates an action with a history of past states and actions, rather than simply with the current state. Various exact and approximate algorithms exist for computing a policy, and the best known are policy-iteration (Howard, 1960) and value-iteration (Bellman, 1962).

## 3.2  DEFINING THE MDP

The states of the MDP for our Recommender system correspond to the states of the predictive (MC) model. This correspondence may not be optimal, especially in light of our experimental results showing the benefits of skipping and clustering. And, in fact, there are methods for learning the state representation (e.g., McCallum, 1996). Nonetheless, such approaches tend to be less accurate initially, and require a large amount of training data (i.e., time online) to gain accuracy. Consequently, we adopt our simple approach.

The actions of the MDP correspond to a recommendation of an item. One can consider multiple recommendations but, to keep our model simple and computationally tractable, we consider single recommendations only. When we recommend an item $x'$, the user can either (1) accept this recommendation, thus transferring from state $\langle x_1, x_2, x_3 \rangle$ into $\langle x_2, x_3, x' \rangle$, or (2) select a non-recommended item. Since item recommendations are generated only after the user has picked a new item, we do not model the case where a user has requested a different set of recommendations without moving to another state.

The rewards in our MDP encode the utilities of selling items (or showing the web pages) as defined by the site. For example, the reward for state $\langle x_1, x_2, x_3 \rangle$ can be the profit generated for the site from the sale of item $x_3$—the last item in the transaction sequence.

The transition function for the MDP model:

$$tr_{MDP}(\langle x_1, x_2, x_3 \rangle, x', \langle x_2, x_3, x'' \rangle)$$

is the probability that the user will select item $x''$ given that item $x'$ is recommended.

Unlike traditional MDP implementations that learn the proper values for the transition function and hence the optimal policy online, our system needs to be fairly accurate when it is first deployed. A for profit e-commerce site would unlikely use a recommender system that generates irrelevant recommendations for a long while, waiting to converge to an optimal policy. We therefore need to initialize the transition function values carefully. We can do so based on our predictive model, making the following assumptions:

- A recommendation increases the probability that a user will buy an item. This probability is proportional to the probability that the user will buy this item in the absence of recommendations. This assumption is made by most CF models dealing with e-commerce sites. We denote the proportionality constant by $\alpha$, where $\alpha > 1$.

- The probability that a user will buy an item that was not recommended is lower than the probability that she will buy it in the absence of recommendations, but still proportional to it. We denote the proportionality constant by $\beta$, where $\beta < 1$.

$$tr_{MDP}(\langle x_1, x_2, x_3 \rangle, x', \langle x_2, x_3, x' \rangle) = \\ \alpha \cdot tr_{MC}(\langle x_1, x_2, x_3 \rangle, \langle x_2, x_3, x' \rangle)$$

$$tr_{MDP}(\langle x_1, x_2, x_3 \rangle, x', \langle x_2, x_3, x'' \rangle) = \\ \beta \cdot tr_{MC}(\langle x_1, x_2, x_3 \rangle, \langle x_2, x_3, x'' \rangle), \ x'' \neq x'$$

Note that $\alpha$ is constant for all initial states and $x'$, whereas $\beta$ is adjusted for each initial state and $x'$ so that the transition probabilities sum to 1.

We note that the MDP can be initialized using any probabilistic predictive algorithm. In particular, to initialize the transition probability from $\langle x_1, x_2, x_3 \rangle$ to $\langle x_2, x_3, x_4 \rangle$, we simply need the predictive algorithm's probability for $x_4$, given that a user purchased $x_1, x_2$, and $x_3$.

### 3.3 SOLVING THE MDP

Given an MDP $\langle S, A, R, tr \rangle$, we need to solve it—that is, determine the best course of action for every possible state. For the domains we have studied, we have found policy iteration (Howard, 1960)—with a few approximations to be described— to be a tractable solution method. In fact, on tests using real data, we have found that policy iteration terminates after a handful of iterations. Fast convergence occurs because we are iterating from the last state to the first and, if there is a non-zero transition probability from $s_i$ to $s_j$, then usually $j > i$ due to the way we build our states and initialize $tr_{MDP}$. Also, we have seen that the computation of the optimal policy is not heavily sensitive to variations in $k$. As $k$ increases, so do the number of states, but the number of positive entries in our transition matrix remains similar. Note that, at most, a state can have as many successors as there are items. When $k$ is small, the number of observed successors for a state is indeed close to the number of items. When $k$ is larger, however, the number of successors decreases considerably.

Although the number of iterations required is small, each iteration requires the computation of the expected rewards for every state, given the current policy. Even though we have reduced the state space by limiting each state to hold the last $k$ transactions, the state space is quite large even for $k = 3$. Thus, the computation of an optimal policy without approximation is extremely time consuming. We reduce run time using the following approximation. The vast majority of states in our models do not correspond to sequences that were observed in our training set. This fact holds because most combinations of items are extremely unlikely. For example, it is unlikely to find adjacent purchases of a science-fiction and a gardening book. In our approximation, we do not compute a policy for a state that was not encountered in our training data. We use the immediate reward of such states as an approximation of their long-term expected value for the purpose of computing the value of a state that appeared in our training data. Of course, if a policy is explicitly required for an unencountered state, we compute its value in the same manner as states that appeared in our training data. This approximation, although risky in general MDPs, is motivated by the fact that in our initial model, the probability of making a transition into an unencountered state is very low.

Once we have a policy, we can use it to generate the recommendation for each state. If more than one recommendation per state is desired, we can (as an approximation) return the $m$ top value recommendations. In theory, we would take into account sets of recommendations, but this approach would be impractical.

Note that the policy calculated is not optimal in the sense that it is based on the transition function initialized from the Markov chain. The policy should be updated online as we now disucss.

### 3.4 UPDATING THE MODEL ONLINE

Once the Recommender system is deployed with its initial model, we can update the model according to actual observations. The simplest approach is to perform off-line updates at fixed time intervals. The main issue is the amount of exploration performed. Without exploration, one cannot update the probability of acceptance of sub-optimal recommendations. Thus, it seems appropriate to select some constant $\epsilon$, such that recommendations whose expected value is $\epsilon$-close to optimal will be allowed—for example, by following a Boltzman distribution with an $\epsilon$ cutoff. The exact value of $\epsilon$ would be determined by the site operators. The price we pay for this conservative exploration policy is that we are not guaranteed convergence to an optimal policy. However, as we have discussed, it is unlikely that significant exploration would be allowed by site operators in practice.

If the system recommends more than one item at each state, another possibility is to allow larger values of $\epsilon$ for the recommendations near the bottom of the recommendation list. For example, the system could always return the best recommendation first, but show items less likely to be purchased as the second and third items on the list.

It is important to note that online learning of the transition function will allow the system to more accurately estimate the probability that a user will decide to purchase item $x$ if item $y$ was recommended. In addition, the online learning procedures should introduce new states and revise the transition function, when an item never before purchased is first purchased.

## 4 TESTING THE PREDICTIVE MODEL

Below, we describe an evaluation of the predictive model described in Section 2.

### 4.1 DATA SETS

For our tests, we used real data from the Israeli online bookstore $Mitos$ (www.mitos.co.il). We used two data sets, one containing user transactions (purchases) and the other containing user browsing paths obtained from web logs. We filtered out items that were bought/visited less than 100 times and users who bought/browsed no more than one item. We were left with 116 items and 10820 users in the transactions data set, and 65 items and 6678 users in the browsing data set. Items that were rarely bought can not be reliably predicted, and our MDP model should learn the transition function for those states online as previously discussed. In our browsing data, no cookies were used by the site. If the same user visited the site with new IP address, then we would treat her as a new user. Also, activity on the same IP address was attributed to a new user whenever there were no requests for two hours. These data sets were randomly split into a training set (0.9 of the users) and a test set (0.1 of the users).

We evaluated predictions as follows. For every user sequence $t_1, t_2, .., t_n$ in the test set, we generated the following test cases: $\langle t_1 \rangle, \langle t_1, t_2 \rangle, ..., \langle t_{n-k}, t_{n-k+1}, ..., t_{n-1} \rangle$. For each case, we then used our various models to determine the probability distribution for $t_i$ given $t_{i-k}, t_{i-k+1}, ..., t_{i-1}$ and ordered the items by this distribution. Finally, we used the $t_i$ actually observed in conjunction with the recommendation list to compute a score for the list.

## 4.2 EVALUATION METRICS

We used two scores: Recommendation Score (RC) (Microsoft, 2002) and Exponential Decay Score (ED) (Breese, Heckerman, & Kadie, 1998) with slight modifications to fit into our sequential domain.

**Recommendation score.** For this measure of accuracy, a recommendation is deemed successful if the observed item $t_i$ is among the top $m$ recommended items ($m$ is varied in the experiments). The score $RC$ is the percentage of cases in which the prediction is successful. A score of 100 means that the recommendation was successful in all cases.

**Exponential Decay Score.** This measure of accuracy is based on the position of the observed $t_i$ on the recommendation list. The underlying assumption is that users are more likely to see a recommendation near the top of the list. In particular, it is assumed that a user will see the $m$th recommendation with probability $p(m) = 2^{-(m-1)/(\alpha-1)}$ ($m \geq 1$). (Note that $\alpha$ is the half-life parameter—the item in the list having probability of 0.5 being seen.) The score is given by

$$100 \cdot \frac{\sum_{c \in C} p(m = pos(t_i|c))}{|C|}$$

where $C$ is the set of all cases, $c = t_{i-k}, t_{i-k+1}, ..., t_{i-1}$ is a case, and $pos(t_i|c)$ is the position of the observed item $t_i$ in the list of recommended items for $c$. We used $\alpha = 5$ in our experiments in order to be consistent with the experiments of Breese et al. (1998) and Zimdars et al. (2001). The relative performance of the models was not sensitive to $\alpha$.

## 4.3 COMPARISON MODELS

**Commerce Server 2002 Predictor** The main model to which we compared our results is the Predictor tool developed by Microsoft as a part of Microsoft Commerce Server 2002, based on the models of (Heckerman et al., 2000). This tool builds dependency- network models in which the local distributions are probabilistic decision trees. We used these models in both a non-sequential and sequential form. These two approaches are described in Heckerman et al. (2000) and Zimdars et al. (2001), respectively. In the non-sequential approach, for every item, a decision tree is built that predicts whether the item will be selected based on whether the remaing items were or were not selected. In the sequential approach, for every item, a decision tree is built that predicts whether the item will be selected next, based on the previous $k$ items that were selected. The predictions are normalized to account for the fact that only one item can be predicted next. Zimdars et al. (2001) also uses a "cache" variable, but preliminary experiments showed it to decrease predictive accuracy. Consequently, we did not use the cache variable in our formal evaluation.

These algorithms appear to be the most competitive among published work. The combined results of Breese et al. (1998) and Heckerman et al. (2000) show that (non-sequential) dependency networks are no less accurate than Bayesian-network or clustering models, and about as accurate as *Correlation*, the most accurate (but computationally expensive) memory-based method. Sarwar, Karypis, Konstan, and Riedl (2000a) apply dimensionality reduction techniques to the user rating matrix, but their approach fails to be consistently more accurate than Correlation. Only the sequential algorithm of Zimdars et al. (2001) is more accurate than the non-sequential dependency network.

We built five sequential models $1 \leq k \leq 5$ for each of the data sets. We refer to the non-sequential Predictor models as Predictor-NS, and to the Predictor models built using the data expansion methods with a history of length $k$ as Predictor-$k$.

**Unordered MCs.** We also evaluated a non-sequential version of our predictive model, where (e.g.) the sequences $\langle x, y, z \rangle$ and $\langle y, z, x \rangle$ are mapped to the same state. Skipping, clustering, and mixture modelling were included as described in Section 2. We call this model UMC (Unordered Markov chain).

## 4.4 VARIATIONS OF THE MC MODEL

In order to measure how each $n$-gram enhancement influenced predictive accuracy, we also evaluated models that excluded some of the enhancements. In reporting our results, we refer to a model that uses skipping and similarity clustering with the terms SK and SM, respectively. In addition, we use numbers to denote which mixture com-ponents are used. Thus, for example, we use MC 123 SK to denote a Markov-chain model learned with three mixture components—a bigram, trigram, and quadram—where each component employs skipping but not clustering.

### 4.5 EXPERIMENTAL RESULTS

Figures 1 and 2 show the exponential decay score for the best models of each type (Markov chain, Unordered Markov chain, Non-Sequential Predictor model, and Sequential Predictor Model). It is important to note that *all* the MC models using skipping, clustering, and mixture modelling yielded better results than *every one of* the Predictor-k models. Thus, to simplify the graphs, only the best models of each class are presented. We see that the sequence-sensitive models are better predictors than those that ignore sequence information. Furthermore, the Markov chain predicts best for both data sets.

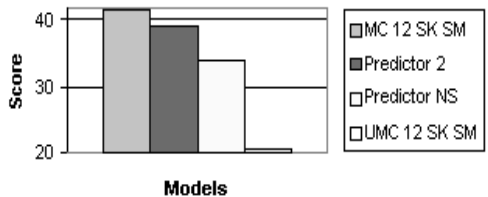

Figure 1: Exponential decay score for different models on transactions data set.

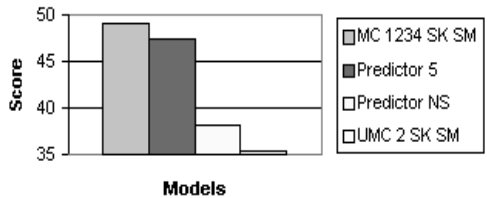

Figure 2: Exponential decay score for different models on browsing data set.

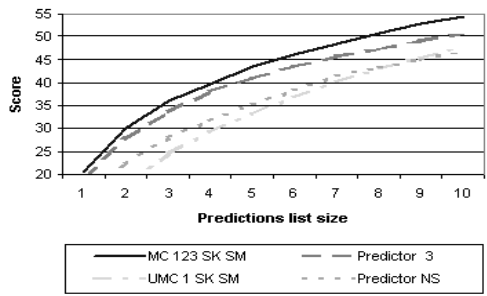

Figure 3: Recommend score for different models on transactions data set.

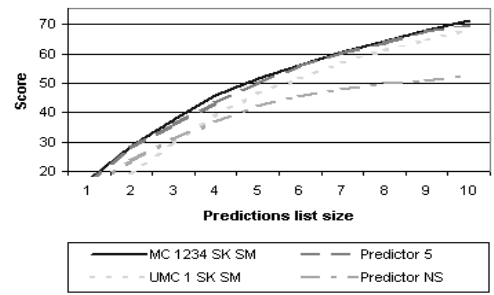

Figure 4: Recommend score for different models on browsing data set.

Figure 3 and Figure 4 show the Recommend score as a function of list length ($m$). Once again, sequential models are superior to non-sequential models, and the Markov-chain models are superior to the Predictor models.

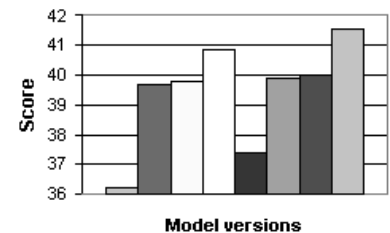

Figure 5: Exponential decay score for different markov chain versions on transactions data set.

Figure 5 and Figure 6 show how different versions of the Markov chain performed under the exponential decay score in both data sets. We see that multi-component models outperform single-component models, and that similarity clustering is beneficial. In contrast, we find that skipping is only beneficial for the transactions data set. Perhaps users tend to follow the the same paths in a rather conservative manner, or site structure does not allow users to "jump ahead". In either case, once recommendations are available in the site (thus changing the site structure), skipping may prove beneficial.

## 5 CONCLUSIONS AND FUTURE WORK

We described a new model for Recommender systems based on an MDP. Our approach makes two main contributions: a conceptual contribution that stems from this new view of Recommender systems, and a technical contribution in the form of a new $n$-gram based Markov-chain pre-dictive model.

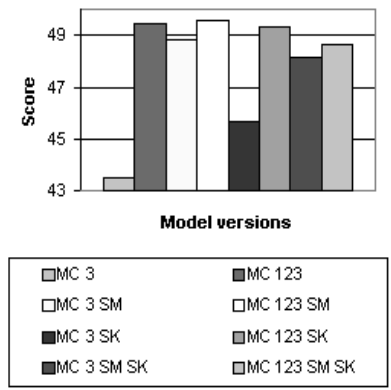

Figure 6: Exponential decay score for different markov chain versions on browsing data set.

Our work also suggests a novel approach to learning MDPs. It takes into account the fact that a deployed MDP-based Recommender system cannot count on standard reinforcement learning techniques for its initialization, as these would lead to initial behavior that would not be tolerated by a site owner. Rather, we use the behavior of customers in the site prior to the deployment of the Recommender system to train a Markov chain that can then be used to initialize the MDP in an informed manner.

Many interesting issues remain for future work. First and foremost, as we noted earlier, our experiments validate only the predictive power of our Markov chain model. Although this limitation is common in evaluations of Recommender systems, it is not an ideal situation. Recommender systems, including ours, need to be evaluated *in situ*.

We are currently in the process of deploying our system in the Israeli online book store $Mitos$, and will soon be able to evaluate the behavior of the system in a live site. We also expect to explore alternatives for initializing the MDP as well as updating the MDP online.

Weaknesses of our predictive (Markov chain) model include the use of *ad hoc* weighting functions for skipping and similarity functions and the use of fixed mixture weights. Although the recommendations that result from our current model are (empirically) useful for ranking items, we have noticed that the probability distributions they produce are not calibrated. Learning the weighting functions and mixture weights from data should improve calibration. In addition, in informal experiments, we have seen evidence that learning case-sensitive mixture weights should improve predictive accuracy.

Our predictive model should also make use of relations between items that can be explicitly specified. For example, most sites that sell items have a large catalog with hierarchical structure such as categories or subjects, a carefully constructed web structure, and item properties such as author name. Finally, our models should incorporate information about users such as age and gender.